\documentclass[preprint,12pt]{elsarticle}

\usepackage{amssymb}

\usepackage{verbatim}
\usepackage{acronym} 

\acrodef{PWM}[PWM]{Pulse Width Modulation}
\acrodef{FSR}[FSR]{Force Sensitive Resistor}
\acrodef{PFM}[PFM]{Pulse Frequency Modulation}
\acrodef{FPGA}[FPGA]{Field Programmable Gate Array}
\acrodef{PID}[PID]{Proportional Integral and Derivative}
\acrodef{DVS}[DVS]{Dynamic Vision Sensor}
\acrodef{VLSI}[VLSI] {Very Large Scale Integration}
\acrodef{AER}[AER]{Address Event Representation}
\acrodef{PCB}[PCB]{Printed Circuit Board}
\acrodef{FSM}[FSM]{Finite State Machine}
\acrodef{LUT}[LUT]{Look-Up Table}
\acrodef{CPG}[CPG] {Central Pattern Generator}
\acrodef{sCPG}[sCPG] {Spiking Central Pattern Generator}
\acrodef{SNN}[SNN]{Spiking Neural Network}
\acrodef{ANN}[ANN]{Artificial Neural Network}
\acrodef{DOF}[DOF]{degrees of freedom}
\acrodef{HDL}[HDL]{Hardware Description Language}
\acrodef{VHDL}[VHDL]{VHSIC Hardware Description Language}
\acrodef{FW}[FW]{Forward}
\acrodef{BW}[BW]{Backward}
\acrodef{LIF}[LIF]{Leaky Integrate-and-Fire}
\acrodef{HBP}[HBP]{Human Brain Project}

\usepackage{ulem}
\usepackage{xcolor}

\graphicspath{{img/}}

\journal{Neural Networks}

\begin{document}

\begin{frontmatter}



\title{Neuromorphic adaptive spiking CPG towards bio-inspired locomotion of legged robots}


\author{Pablo~Lopez-Osorio\fnref{label1}}
\ead{pablo.osorio@uca.es}
\author{Alberto~Patino-Saucedo\fnref{label2}}
\author{Juan~P.~Dominguez-Morales\fnref{label3}}
\author{Horacio~Rostro-Gonzalez\fnref{label2}}
\author{Fernando~Perez-Pe\~{n}a\fnref{label1}}
\ead{fernandoperez.pena@uca.es}

\address[label1]{School of Engineering, Universidad de C\'{a}diz, Spain.}
\address[label2]{Department of Electronics, DICIS-University of Guanajuato, Mexico}
\address[label3]{Robotics and Technology of Computers Lab. Universidad de Sevilla, Spain.}

\begin{abstract}

In recent years, locomotion mechanisms exhibited by vertebrate animals have been the inspiration for the improvement in the performance of robotic systems. These mechanisms include the adaptability of their locomotion to any change registered in the environment through their biological sensors. In this regard, we aim to replicate such kind of adaptability in legged robots through a \Acl{sCPG}. This \Acl{sCPG} generates different locomotion (rhythmic) patterns which are driven by an external stimulus, that is, the output of a \Acl{FSR} connected to the robot to provide feedback. The \Acl{sCPG} consists of a network of five populations of \Acl{LIF} neurons designed with a specific topology in such a way that the rhythmic patterns can be generated and driven by the aforementioned external stimulus. Therefore, the locomotion of the end robotic platform (any-legged robot) can be adapted to the terrain by using any sensor as input. The \Acl{sCPG} with adaptive learning has been numerically validated at software and hardware level, using the Brian~2 simulator and the SpiNNaker neuromorphic platform for the latest.
In particular, our experiments clearly show an adaptation in the oscillation frequencies between the spikes produced in the populations of the \Acl{sCPG} while the input stimulus varies. To validate the robustness and adaptability of the \Acl{sCPG}, we have performed several tests by variating the output of the sensor. These experiments were carried out in Brian~2 and SpiNNaker; both implementations showed a similar behavior with a Pearson correlation coefficient of 0.905.

\end{abstract}


\end{frontmatter}


\section{Introduction}
\label{Sec_Intro}
It is well known that, in biology, rhythmic locomotion is produced by a neural structure called \ac{CPG} \cite{ijspeert2008central}. This structure is located at the spinal cord and it usually comprises of two neural populations which produce an alternating output of spikes. Eventually, these output spikes are used to activate the muscles fibers.  

This approach of using \acp{CPG} can be borrowed to create locomotion in robotics. There are several possibilities to implement a \ac{CPG}: using coupled-oscillators, using \acp{ANN} or using \acp{SNN}. The closest biological implementation is to use an \ac{SNN}. These networks are based on neuron models and synaptic connections that implement biological features. The field of research called neuromorphic engineering aims to implement these networks on electronics, mimicking the way living beings have solved complex problems by using both analog and digital circuits. The neuromorphic robotics field puts together both the neuromorphic engineering and the roboticists communities \cite{krichmar2011neuromorphic}. The use of neural structures made of spiking neurons, coming from the neuromorphic engineering field, within robotics results in the need for less resources, less power consumption and a simplification of the algorithms \cite{indiveri2011neuromorphic}.

One of this neural structures is the \ac{CPG}. This structure generates a rhythmic pattern at its output, which can be used within robotics to generate locomotion. Thus, a \ac{CPG} creates gaits that are suitable to use within a robotic platform. These structures can generate a very stable pattern even without sensory information or brain activity \cite{Vogelstein2008}.

As briefly shown, there is a growing community of researchers that are exploring the possibility of using a \ac{CPG} to create locomotion in robots \cite{ijspeert2008central}. Most of the previous works in the literature present an open-loop \ac{CPG} which does not include any sensory information: in \cite{Donati2014}, the actuation of a lamprey-like robot is done by using an open-loop \ac{CPG} and neuromorphic hardware. Then, in \cite{rostro2015cpg} and \cite{Brayan_2017}, the authors proposed a \ac{CPG} implemented on an \ac{FPGA} and the SpiNNaker platform \cite{furber2014spinnaker}; these three works do not offer the possibility of changing the originally produced pattern in real time. However, they showed that implementing \acp{CPG} using spiking neurons uses less power. There is a more recent paper which allows both real time functioning and pattern variation but without including any sensory information \cite{gutierrez2020neuropod}. In \cite{Polykretis2020}, the open-loop \ac{CPG} is implemented on Loihi \cite{davies2018loihi} using an astrocytic network, producing two different gaits with 24 motor neurons.
However, the most recent work \cite{Strohmer2020}, proposes the implementation of a \ac{CPG} with the possibility of changing the amplitude, frequency and phase online without any sensory input required. The authors also pointed out that the architecture should include sensory feedback to modify the behavior of the \ac{CPG}. That is the objective presented in this paper. We propose an \ac{SNN} that, using the sensory information, adapts the behavior of the \ac{CPG}.   
Regarding works that include sensory information, in \cite{Sartoretti2018}, the \ac{CPG} is built using coupled-oscillators instead of spiking neurons and the feedback to the \ac{CPG} is included into the control loop of the equations of the \ac{CPG}. In \cite{Spaeth2020}, 12~simulated neurons modulated by sensory feedback are used to build the \ac{CPG}. Instead of what we propose, an adaptation \ac{SNN}, they achieve different gaits by either moving the location or increasing the number of neural structures. The neuron model used is Izhikevich instead of the \ac{LIF} model proposed in this work to reduce the computational resources used. The sensory information is used to adapt the time duration of each phase of the frequency switching to enable the actuators to reach the commanded position. Finally, in \cite{gutierrez2019live}, a neuromorphic sensor has been used to select which predefined gait of the \ac{CPG} should be activated. In most of these works, the use of an external input to the \ac{CPG} changes the performance of it. This performance can be defined as a neuromodulation. It has been shown that this modulation is essential to alter the behavior of a neural structure by modifying the synaptic connections \cite{harris2011neuromodulation}. 

Finally, there are works where the main focus is on the learning process of the robotic platform: in \cite{Lele2020}, the authors proposed a rewarding-learning process to teach a hexapod robot how to walk without any previous knowledge. A couple of sensors (a standard camera and a gyroscope) are used to provide the rewarding signal to the neural network based on a \ac{CPG}. Although they used a digital version model neuron of the \ac{LIF}, they do not use neuromorphic hardware to implemented the neural architecture; a Raspberry Pi is used instead. A similar approach based on reinforcement learning, but without using spiking neurons, was proposed to improve the locomotion of the NAO robot \cite{li2013humanoids}. 
Another approach is used in \cite{Ting}, where the authors have two hexapod robots: an expert and a student. The student learns or imitates the gait of the expert by using a one-layer feedforward \ac{SNN} and a \ac{DVS} camera as input. However, the possibility for the robot to adapt to its environment is not implemented.  


The objective of this paper is to design and deploy a spiking architecture that makes the interaction of a spiking \ac{CPG} with its environment possible. An external agent. i.e. a \ac{FSR}, is introduced as the feedback stimulus to the network. This agent can modify the gait generated by the \ac{CPG}. Therefore, the locomotion of the end robotic platform (any legged robot) can be adapted to the terrain. Furthermore, the spiking network presented in this paper allows the introduction of the feedback sensory information on the loop of the \ac{CPG} to provide adaptation. This adaptation \ac{SNN} could be used with any sensor as stimulus. 


The rest of the paper is structured as follows: section~\ref{Sec_Materials} introduces the materials used in this work, including the simulator and hardware used. The implemented methods are described in \ref{Sec_Methods}, together with the \ac{SNN} model. Then, the results obtained are presented and discussed in section~\ref{Sec_Results&Discussion}. This section is divided into two different subsections: first, the experiments run using the software simulator and then, the same experiments run on on the hardware platform. Finally, the conclussions are presented. 
\section{Materials and Methods}
\subsection{Materials}
\label{Sec_Materials}
This section describes both the software and hardware used to perform the experiments. 
\subsubsection{Brian~2}
\label{subsubsec_brian}

Brian~2 \cite{stimberg2019brian} is a neural simulator for \acp{SNN} written in Python programming language. Thus, it is a cross-platform which is available in different operating systems. In contrast to other \ac{SNN} simulators such as NEURON \cite{carnevale2006neuron} or PyNN \cite{davison2009pynn}, Brian~2 is highly flexible and it is easily adaptable with new non-standard neuron models and synapses. Brian~2 can be used to model and simulate complex problems faced by neuroscientists, as well as giving faster and more robust results before implementing the solution on a hardware platform.

\subsubsection{SpiNNaker}
\label{subsubsec_spinnaker}

SpiNNaker \cite{furber2012overview, furber2014spinnaker, furber2020spinnakerbook} is a massively parallel, multi-core computing system designed by the Advanced Processor Technologies (APT) Research Group from the University of Manchester. It was designed under the \ac{HBP} \cite{markram2012human} for simulating parts of the brain by using \acp{SNN}. SpiNNaker machines consist of SpiNNaker chips, which have eighteen 200-Hz ARM968 processor cores each \cite{painkras2013spinnaker}. This allows an asynchronous communication infrastructure for sending short packages (each of them representing a particular neuron firing) \cite{plana2007gals} identified using \ac{AER} \cite{mahowald1992vlsi}. Different SpiNNaker machines were built and commercialized, including SpiNN-3 and SpiNN-5, with 4 and 48 SpiNNaker chips each, respectively. They also include the spinnlinks \cite{plana2020spinnlink}, which allow real-time input/output interfacing with neuromorphic sensors and other neuromorphic platforms such as \acp{FPGA} \cite{yousefzadeh2017multiple, dominguez2016multilayer, schoepe2019neuromorphic}. A PyNN-based \cite{davison2009pynn} software package called sPyNNaker \cite{rhodes2018spynnaker} can be used to design and implement \acp{SNN} on these machines. The recently built million-core machine is at the School of Computer Science at the University of Manchester that can be used through the \ac{HBP} portal. In this work, a SpiNN-5 machine was used to run the simulations proposed.

\subsection{Methodology}
\label{Sec_Methods}
We simulated our \ac{sCPG} model on a standard computer using the Brian~2 Simulator to characterize the network dynamics, to analyse the number of neurons per population needed and to adjust the network parameters. The simulation results guided the subsequent neuromorphic implementation using the SpiNNaker platform. 
\subsubsection{Neuron Model}
\label{subsubsec:neuron_model}
The \ac{LIF} model is used to implement the neuron on both the software simulator and the SpiNNaker hardware platform \cite{gerstner2002}. The model is defined within the equations (\ref{lif_eq_1}) and (\ref{lif_eq_2}).
\begin{equation}
\label{lif_eq_1}
\tau_{m} \frac{dV}{dt} = -(V-V_{r})+RI(t) 
\end{equation}
\begin{equation}
\label{lif_eq_2}
if\: V(t) = V_{th}\quad then\quad \lim_{\delta \rightarrow 0; \delta > 0} V(t+\delta) = V_{r}
\end{equation}
where $V$ is the membrane potential of the neuron, $R$ represents the resistance of the membrane, $\tau_m$ the time constant of the neuron, $V_{r}$ the resting potential, $V_{th}$ the threshold and $I(t)$ is the stimulus.

\subsubsection{Network Model}
\label{subsubsec:network_model}
\begin{figure}[h]
\centering\includegraphics[width=0.75\linewidth]{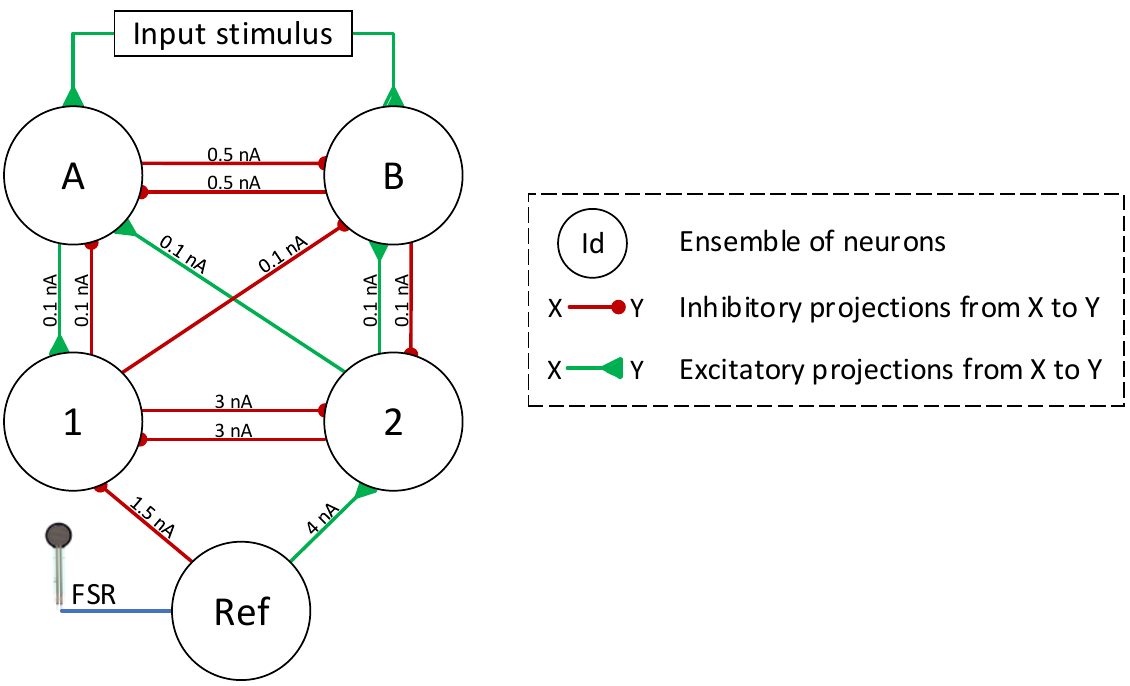}
\caption{Diagram of the proposed spiking neural network architecture.}
\label{fig:SNN_architecture}
\end{figure}

The \ac{SNN} model depicted in Figure~\ref{fig:SNN_architecture} was designed based on the neuron model presented in section~\ref{subsubsec:neuron_model}. The main objective of the proposed model is to generate a constant oscillation between the spikes produced in ensembles~A and~B, whose frequency varies depending on the value read from the \ac{FSR} sensor. Thus, the proposed \ac{CPG} is able to automatically adapt its behavior depending on the input stimulus.

The \ac{SNN} consists of different parts which are described next. It is important to note that each of the ensembles (also called populations) have the same number of neurons. The number of neurons in each population was set based on different experiments, which are presented in section \ref{Sec_Results&Discussion}.

The main block of the architecture is the so-called $CPG_{AB}$, which consists of the populations~A and~B represented in Figure~\ref{fig:SNN_architecture}. These populations are self-excited and self-inhibited with a probability of 25\% and 75\% and weights of 4~nA and 1.5~nA, respectively. Moreover, a 75\% probability of having inhibitory synapses between neurons from the two aforementioned populations is also present. Thus, when one of the populations is producing spikes, the opposite is inhibited and, thus, generating the desired oscillation pattern. Populations~A and~B are injected with a constant external current in order to start generating the oscillation. Therefore, for these two populations equation~\ref{equation_CPG_AB} is used instead of equation~\ref{lif_eq_1}.

\begin{equation}
\label{equation_CPG_AB}
\frac{dV}{dt} = \frac{Vr-V+R(Iexc-Iinh+Ist)}{\tau_m}
\end{equation}
Where $I_{st}$ is the current injected to neurons in populations~A and~B. This value was set to 2.2~nA, which is sufficient for producing the desired rhythmic pattern.

Furthermore, populations~1 and~2 ($CPG_{12}$) both have the same number of neurons as those in $CPG_{AB}$, and are also interconnected in the same way. The projections between $CPG_{12}$ and $CPG_{AB}$ are depicted in Figure~\ref{fig:SNN_architecture}, and follow the same aforementioned probabilities (25\% for excitatory and 75\% for inhibitory projections). The weights between the different populations of the proposed model are specified in the same figure.

Finally, the reference population ($Ref$) was implemented as a Poisson distribution with variable frequency. In contrast to the populations from $CPG_{AB}$ and $CPG_{12}$, the number of neurons in $Ref$ was set to 50. These neurons are connected to populations~1 and~2 following the scheme presented in Figure ~\ref{fig:SNN_architecture}. This number of neurons was set to 50 since it was the optimum for producing biologically-plausible spiking rates in the population, with maximum and minimum frequencies close to the biological counterpart \cite{tripathy2014neuroelectro}. 
The spiking rate of the Poisson distribution depends on the values obtained from the \ac{FSR} sensor used as input to population $Ref$. This sensor should be placed at the end of the leg of the robot. This sensor provides values between 0 and 5~V. Since the spinal cord ventral horn motor neuron alpha, which is the biological neuron taken as reference, has a spike rate between 10 and 171~Hz \cite{tripathy2014neuroelectro}, a linear regression was established in order to  match the frequency of the Poisson distribution with the voltage value read from the sensor. Equation~\ref{eq_linearRegression}, where $V_s$ is the voltage value provided by the sensor, shows this relation to provide the desired rates. 

\begin{equation}
\label{eq_linearRegression}
f_{Ref}=\frac{280V_s-950}{3}
\end{equation}


\begin{table}[ht]
\begin{center}
\begin{tabular}{|c|c|}
\hline
\textbf{Parameter}         &\textbf{Value} \\ \hline
\textbf{${u_{reset}}$}                   & -55.0 mV                      \\ \hline
\textbf{$ {u_{rest}}$}                    & -55.0 mV                      \\ \hline
\textbf{${u_{th}}$}                     & 15.0 mV                    \\ \hline
\textbf{$ {\tau_{m}}$}                 & 6.0 ms                    \\
\hline
\textbf{$ {\tau_{syn_e}}$}                 & 5.0 ms                    \\
\hline
\textbf{$ {\tau_{syn_i}}$}                 & 8.75 ms                    \\ \hline
\textbf{$ {c_{m}}$}                    & 0.1875 nF                  \\ \hline
\textbf{$ {\Delta_{t}}$}                  & 1.0 ms                      \\ \hline
\textbf{${I_{bias}}$}                  & 2.2 mA                      \\ \hline
\end{tabular}
\end{center}
 \caption{\label{Table:NeuParams} Neuron parameters for the proposed \ac{CPG} in both the Brian~2 simulator and the SpiNNaker hardware platform.}
\end{table}

Based on these three blocks ($CPG_{AB}$, $CPG_{12}$ and $Ref$) and the connections between them, the proposed behavior explained at the begining of this section was achieved. Therefore, following Fig.~\ref{fig:SNN_architecture}, different scenarios can be analyzed. In the case where the spike rate of population $Ref$ is greater than the oscillation frequency of $CPG_{AB}$, population~1 will be inhibited and population~2 will be excited. Since population~1 is excited from A, but the number of spikes is lower than the ones that are inhibiting the same population from $Ref$, population 1 will have very low activity. The opposite occurs in population~2, which will be inhibited from B but excited from $Ref$ in a stronger way and, thus, having considerable more activity than population~1. The activity from population~2 will excite $CPG_{AB}$, increasing its oscillation frequency. Conversely, the opposite happens when the spike rate of $Ref$ is lower than the oscillation frequency of $CPG_{AB}$, where population~1 will have more activity than 2 and, therefore, inhibit $CPG_{AB}$ in order to reduce its frequency. As a result, the proposed network is able to adapt the frequency of the \ac{CPG} based on an input stimulus. This model was simulated in Brian~2 and emulated using SpiNNaker, and the results can be seen in section~\ref{Sec_Results&Discussion}.

\section{Results and Discussion}
\label{Sec_Results&Discussion}

\subsection{Brian~2 simulations}
\begin{figure}[t]
    \centering
    \includegraphics[width=.8\textwidth, keepaspectratio]{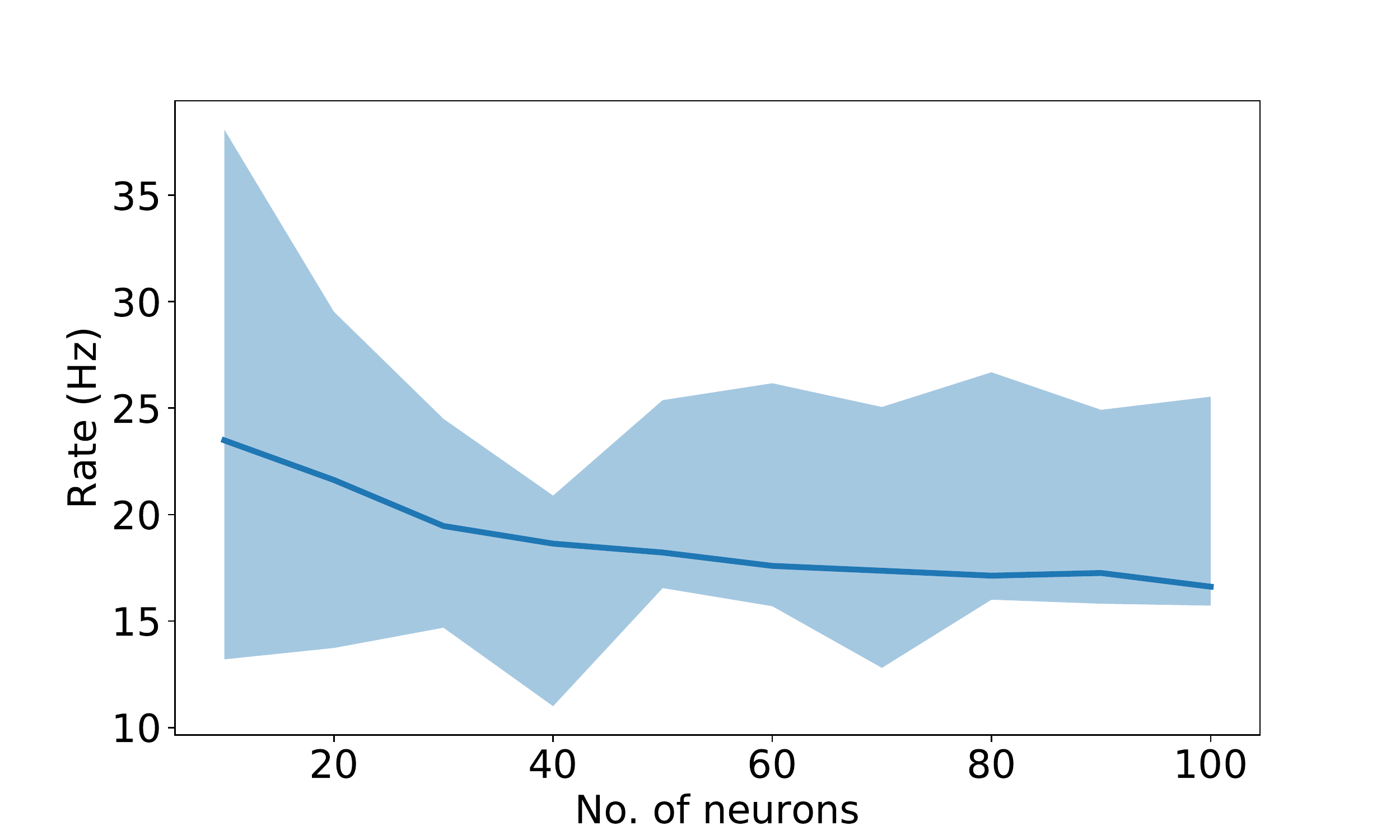}
    \caption{Average, maximum and minimum rate values obtained for each number of neurons per population. The trace shows the mean, and, its shadow, the maximum and minimun rate of all the one thousand simulations performed.}
    \label{fig:IndexPopulations}
\end{figure}

The first experiment was performed to determine the number of neurons per population of the \ac{CPG} that was needed to achieve a stable rate value along the simulations. The parameters used for all the experiments are shown in Table~\ref{Table:NeuParams}. Figure~\ref{fig:IndexPopulations} shows the maximum, minimun and mean values obtained for the rate of the simulations performed. A thousand simulations per number of neurons were run. As the number of neurons increased, the rate achieved is more stable and the standard deviation becomes lower. Starting from 40 neurons per population in the $CPG_{AB}$, the standard deviation is less than 1.5 and the bahavior is more stable. A hundred neurons per population shows the most stable output rate with the least deviation. 
Once the number of neurons per population was fixed, to verify if the architecture presented in Section~\ref{Sec_Methods} behaved as expected, the operation of the $CPG_{AB}$ in isolation was analyzed. This experiment ensured that it was able to produce a constant oscillation. Then, the operation of the same \ac{CPG} was examined once interconnected with the $CPG_{12}$, performing tests with different stimuli to analyze the results obtained. Finally, the entire architecture was connected and analyzed in different scenarios based on the external input from the \ac{FSR} sensor.

\subsubsection{$CPG_{AB}$ analysis}
\label{subsubsec:CPGAB}

The topology of the $CPG_{AB}$ can be seen in Figure ~\ref{fig:SNN_architecture}, where green arrows denote excitatory connections and red arrows denote inhibitory connections. As mentioned in section~\ref{subsubsec:network_model}, $I_{St}$ is a constant current injected to all neurons in populations~A and~B, with a fixed value of 2.2~nA.

\begin{figure}[ht]
    \centering
    \includegraphics[width=\textwidth, keepaspectratio]{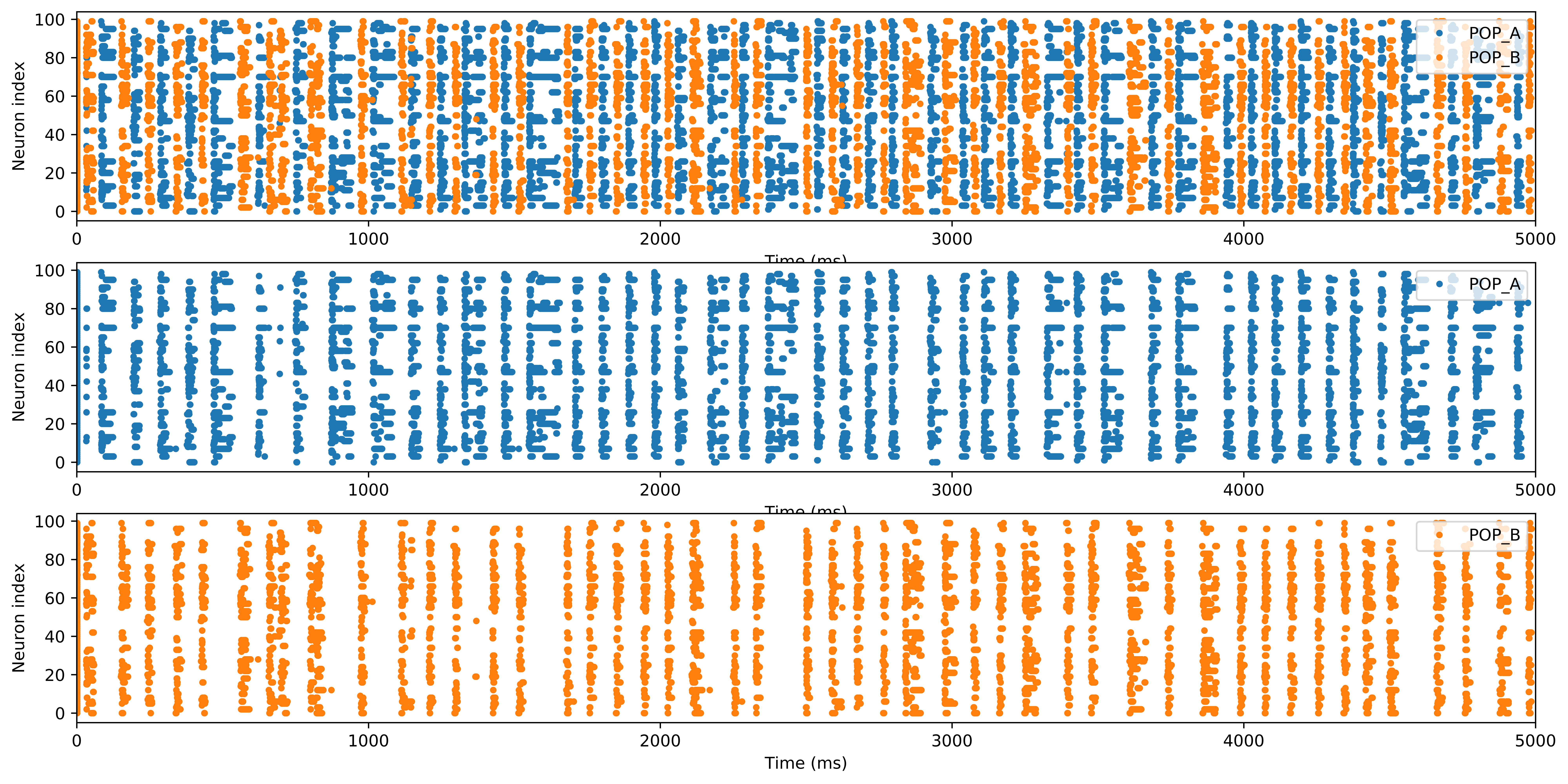}
    \caption{Simulation of the $CPG_{AB}$ with a $I_{St}$ value of 10~nA. Increased oscillation frequency may be observed along with an increase in the amount of noise when compared to Figure \ref{fig:CPGAB22na}.}
    \label{fig:CPGAB10na}
\end{figure}

\begin{figure}[h!]
    \centering
    \includegraphics[width=\textwidth, keepaspectratio]{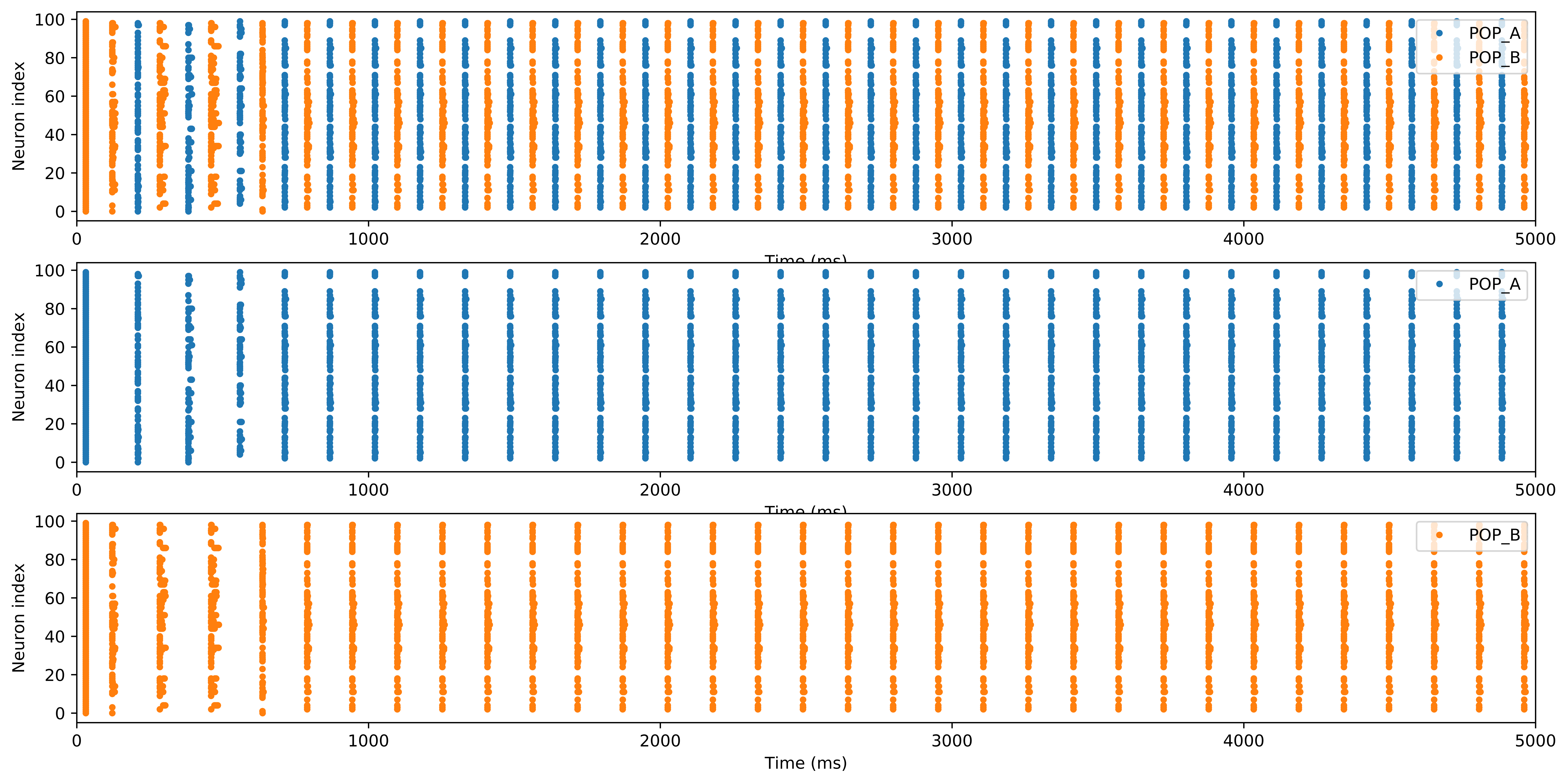}   
    \caption{Simulation of the $CPG_{AB}$ with a $I_{St}$ value of 2.2~nA. A lower oscillation frequency and almost total noise removal can be observed when compared to Figure \ref{fig:CPGAB10na}.}    
    \label{fig:CPGAB22na}
\end{figure}

This current is the minimum value required to produce the oscillatory pattern in the \ac{CPG}. While the proposed topology can work with higher values of $I_{St}$, this generates a higher frequency of oscillation and a noticeable increase in the noise introduced in the simulation (see figures~\ref{fig:CPGAB10na} and~\ref{fig:CPGAB22na}). Thus, this value was set to 2.2~nA in order to be able to more easily appreciate the effect of feedback on the \ac{CPG}. The results of the simulation performed are shown in Figure~\ref{fig:CPGAB22na}, where a frequency of, approximately, 5.8~Hz was obtained for each population, while the total frequency of the \ac{CPG} was 11.6~Hz.

\begin{figure}[t]
    \centering
    \includegraphics[width=\textwidth, keepaspectratio]{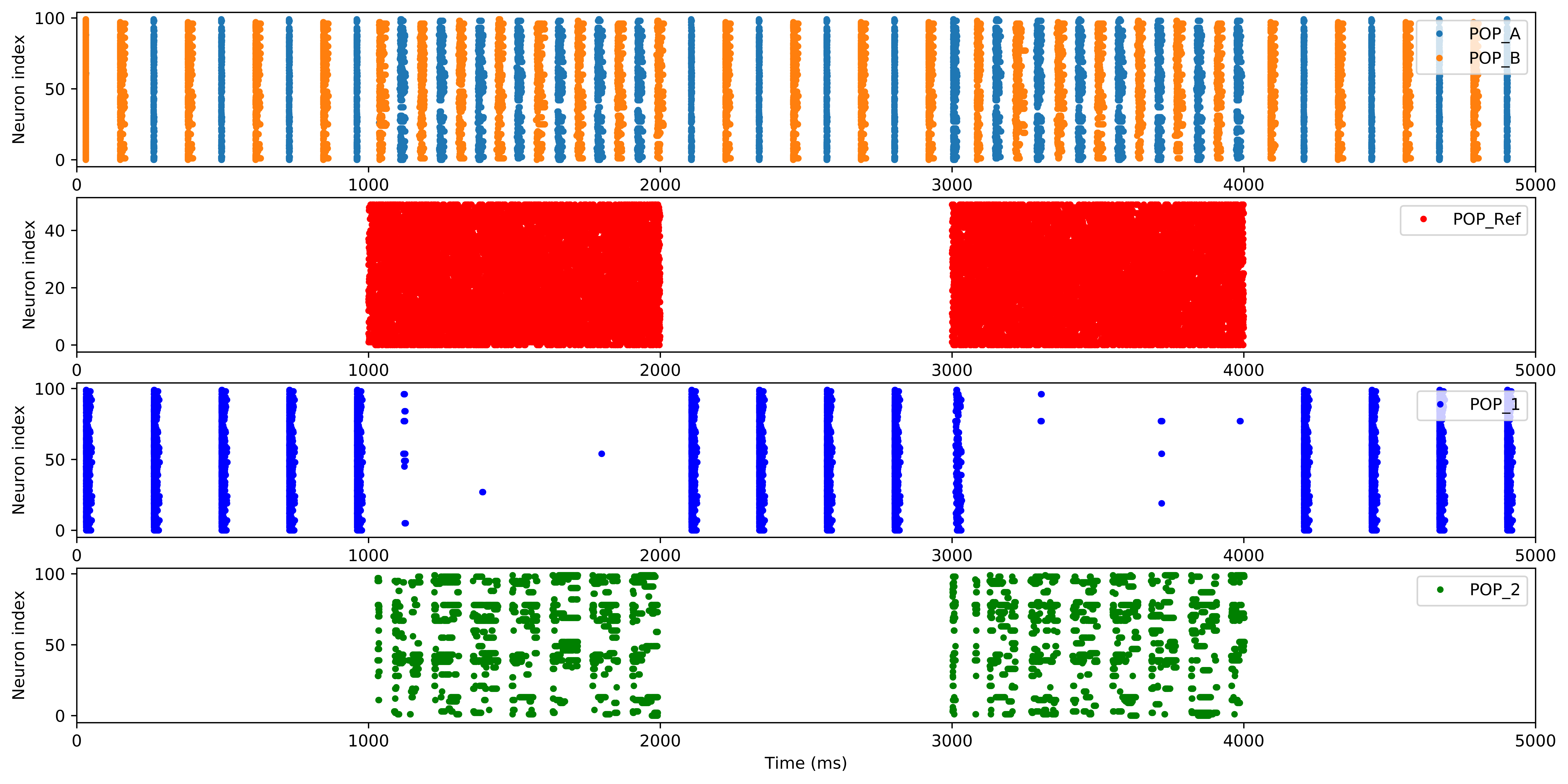}   
    \caption{Results of the simulation of the proposed \ac{SNN} model when having a 5~V input from the \ac{FSR} sensor for 1000~ms between $t=1000 ms$ and $t=2000 ms$ and between $t=3000 ms$ and $t=4000 ms$ (generating a Poisson distribution of 171~Hz in $Ref$). For the rest of the simulation time, the sensor's output was set to 0~V.}  
   \label{fig:CPGS-0-5V}
\end{figure}

\begin{figure}[ht]
    \centering
    \includegraphics[width=\textwidth, keepaspectratio]{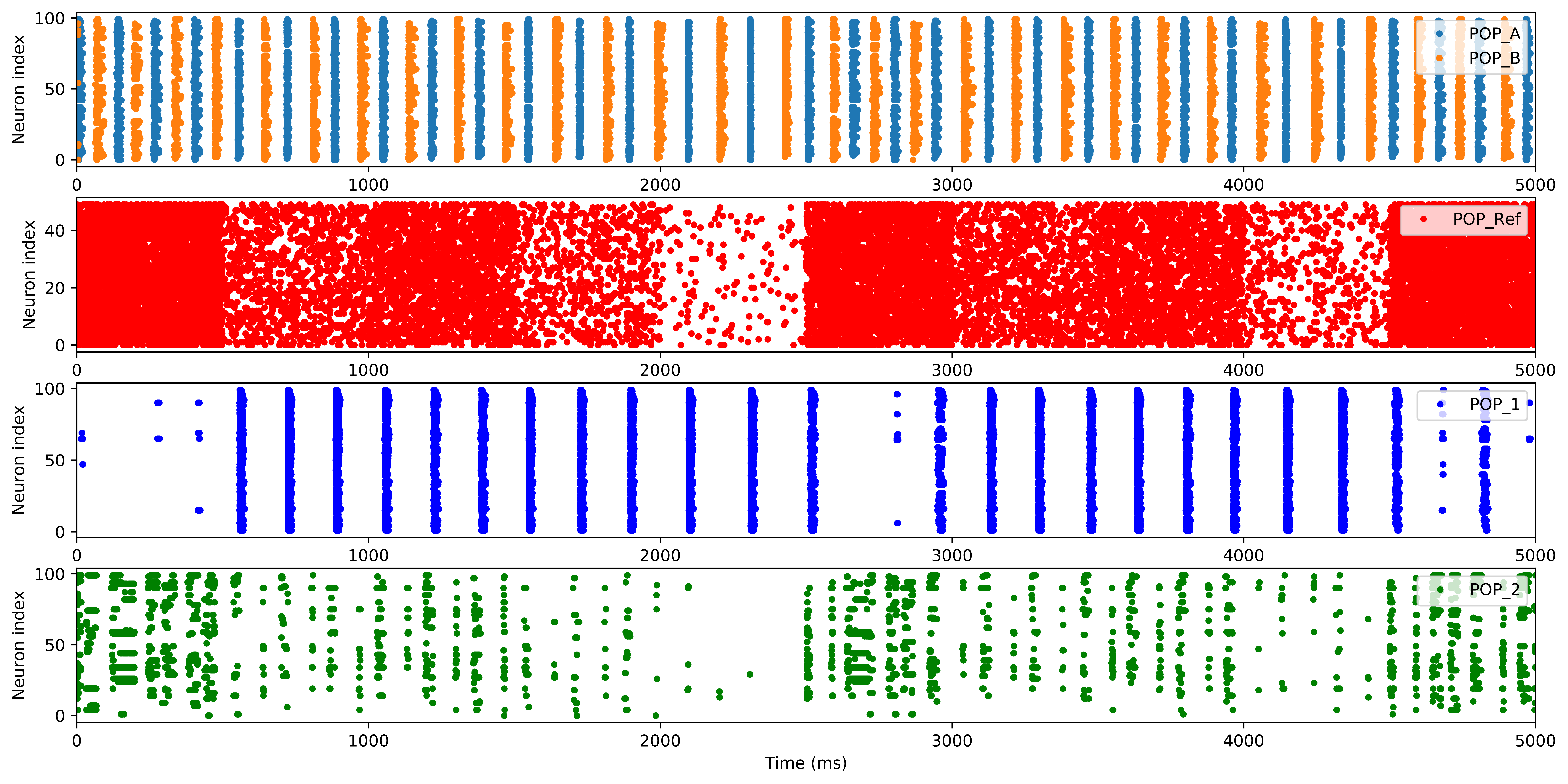}   
    \caption{Results obtained when simulating random voltage readings from the \ac{FSR} sensor.}    
   \label{fig:CPGSal1}
\end{figure}

\begin{figure}[ht]
    \centering
    \includegraphics[width=\textwidth, keepaspectratio]{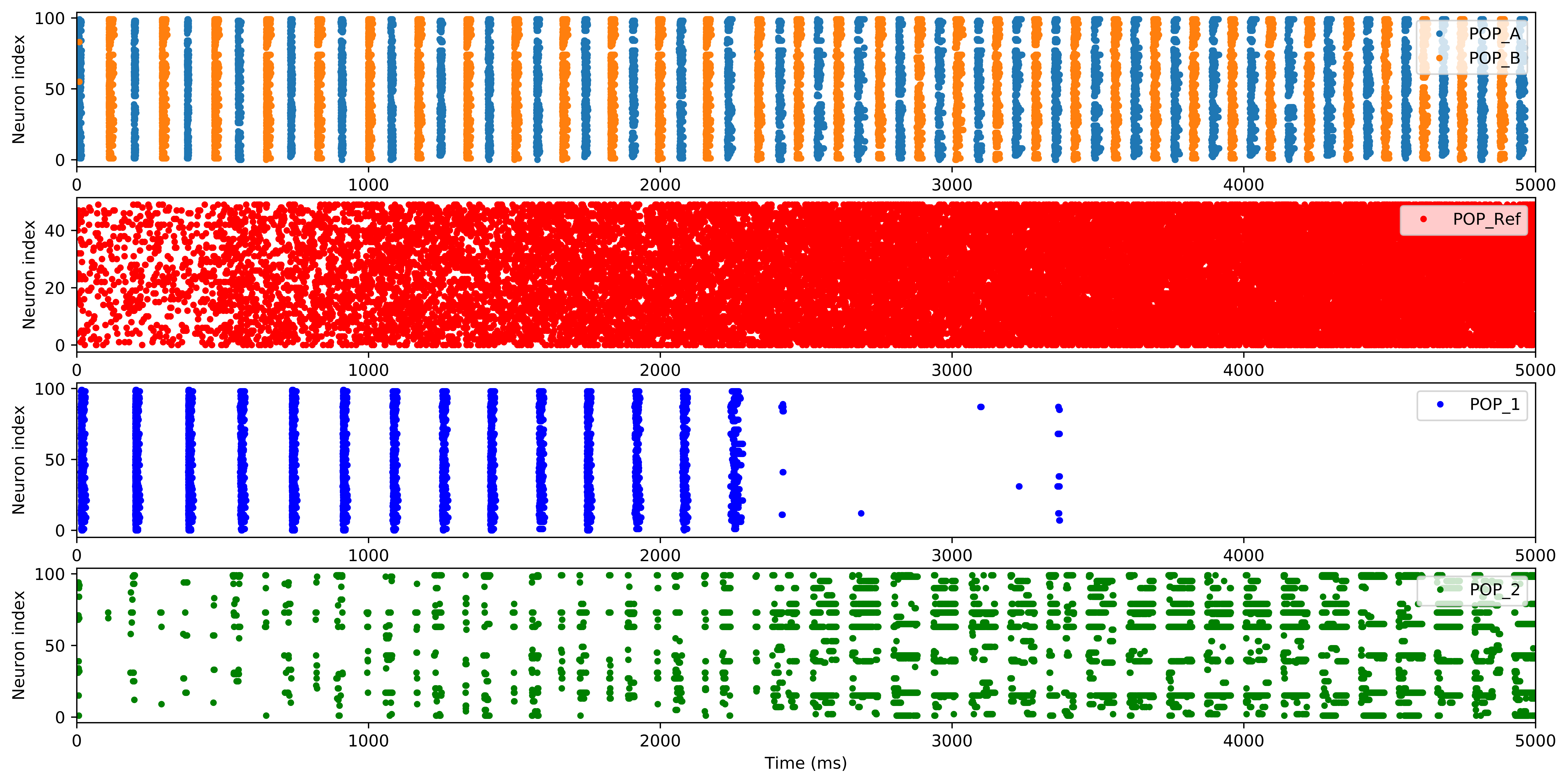}   
    \caption{Results obtained when simulating a continuous increase in the frequency of $Ref$.}    
   \label{fig:CPGSal2}
\end{figure}

\subsubsection{Analysis of the effect of the sensor when used as input to the SNN}
In order to study the robustness of the network against sudden changes in the oscillation frequency, an experiment where the values read from the sensor were alternating between maximum and minimum voltage peaks was performed. Initially, a 5~V input was simulated in 1~s and 2~s, both with a duration of 1~s. This made neurons in population $Ref$ to fire at a frequency of, approximately, 171~Hz during this period. Before 1~s, between 2~s and 3~s, and after 4~s the sensor readings corresponded to 0~V. Figure \ref{fig:CPGS-0-5V} shows the results of this simulation. It can be observed that, at time zero, since no information was being received from the sensor, $Ref$ had no activity. Therefore, population 1 was excited by $CPG_{AB}$ and population~2 was inhibited. In turn, population 1 slightly inhibited $CPG_{AB}$. At 1~s, $Ref$ started firing at a frequency of approximately 171~Hz, exciting population~2 and inhibiting population~1. During this period, the former excited $CPG_{AB}$, increasing its oscillation frequency. After 2~s, population~1 started dominating population~2 again, slightly inhibiting $CPG_{AB}$ again. Exactly the same behavior can be seen again starting at 3~s. In this figure it can be observed how the frequency of $CPG_{AB}$ increased considerably between 1~s and 2~s and between 3~s and 4~s, obtaining minimum frequencies of 8~Hz and maximum frequencies of 15~Hz. 

Finally, different simulations of more realistic cases were performed. Initially, 10 random voltage values were used in order to simulate different readings from the \ac{FSR} sensor. These values were updated every 500~ms. Specifically, the frequency values for Poisson distribution in $Ref$ were (171, 40, 80, 30, 5, 130, 50, 76, 20, 150)~Hz. In Figure~\ref{fig:CPGSal1} it can be seen how $CPG_{AB}$ adapts its oscillation frequency based on the inputs stimuli, obtaining maximum and minimum frequency peaks of 15~Hz and 8~Hz, respectively.

After this, a constant increase in the values of the readings from the sensor was simulated. In particular, increases in steps of 20~Hz per 500~ms were introduced in the frequency of $Ref$. To check the performance limits of the $CPG_{AB}$, the last injected frequency value exceeded up to 17\% of the maximum theoretical value of 171~Hz. Figure \ref{fig:CPGSal2} shows the results of this experiment.

As can be seen in the figure, although there is a significant increase in the amount of noise, the oscillation frequency of $CPG_{AB}$ does not increase, achieving a maximum frequency of 15~Hz.

\subsection{Running the model on SpiNNaker}

The proposed \ac{sCPG} model was tested in the SpiNNaker neuromorphic hardware. The neuronal model is defined in sPyNNaker \cite{rhodes2018spynnaker}, a PyNN-based software interface that allows a quick prototyping and implementation of spiking neural networks in the SpiNNaker platform.

The spiking neuron model is the \ac{LIF} with fixed threshold and decaying-exponential post-synaptic current, whose parameters are given in table \ref{Table:NeuParams}. These parameters were chosen so as to emulate the neuron dynamics of the simulations in Brian~2.

In order to show that the model achieves a good performance in SpiNNaker, we performed three tests: first, by verifying that the constant oscillations of the $CPG_{AB}$ were observed and matched the rates presented in Brian~2. Then, by implementing the whole \ac{sCPG} with increasing rates of the input sensor represented by the reference population. Finally, by testing the \ac{sCPG} under random stimuli. 

\subsubsection{$CPG_{AB}$ implementation}

\begin{figure}[ht]
    \centering
    \includegraphics[width=0.95\textwidth, keepaspectratio]{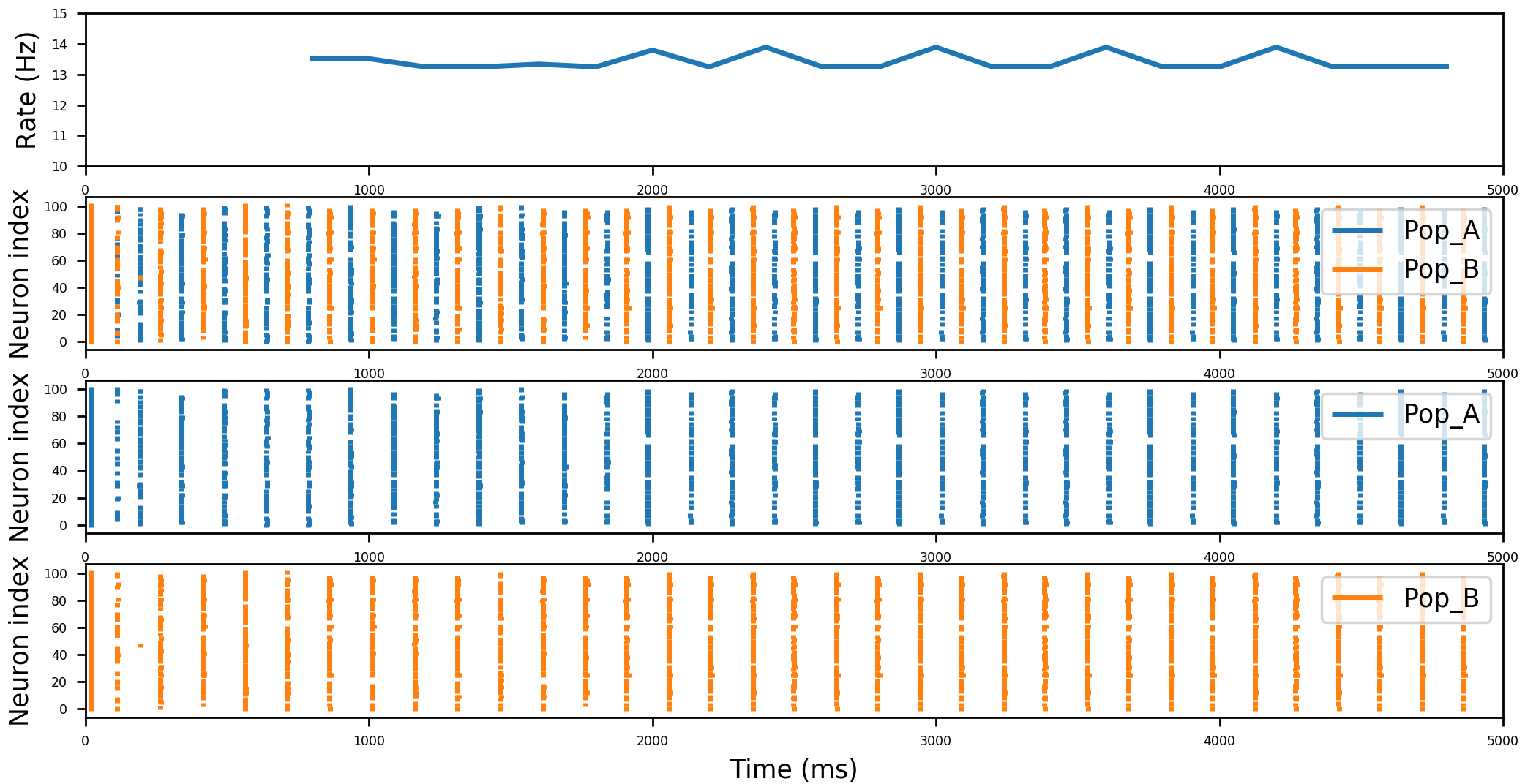}   
    \caption{SpiNNaker simulation of the $CPG_{AB}$.}    
   \label{fig:CPGAB_spinn}
\end{figure}

\begin{figure}[h!]
    \centering
    \includegraphics[width=0.95\textwidth, keepaspectratio]{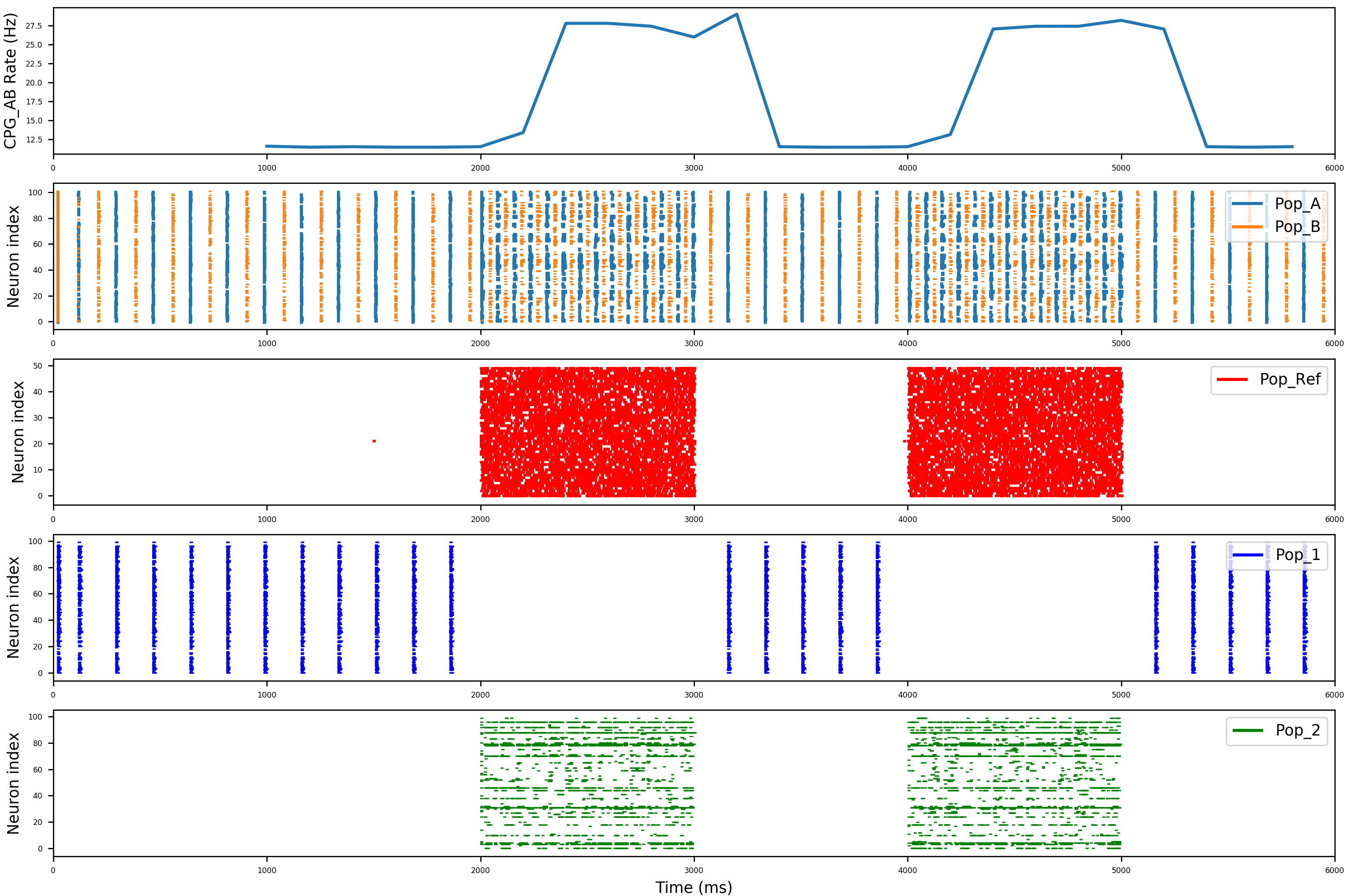}   
    \caption{SpiNNaker simulation of the $CPG$ for extreme values of the stimulus spiking rate (0.1~Hz and 171~Hz).}    
   \label{fig:CPGAB_barrido1}
\end{figure}

For the implementation of the CPG in Spinnaker (see Figure~\ref{fig:CPGAB_spinn}) 100 neurons were used, each with a constant current $I_{St}$ equal to 2.2~nA, like in the equivalent Brian experiment. The measured rate of the oscillatory pattern of populations~A and~B was 11.62~Hz, which matches the rate measured in the Brian simulation. Although some spikes were lost in the raster plot compared to Brian, which can be attributed to limited support of floating-point calculations in sPyNNaker, the pattern appears consistent and with little noise.

\begin{figure}[t]
    \centering
    \includegraphics[width=0.95\textwidth, keepaspectratio]{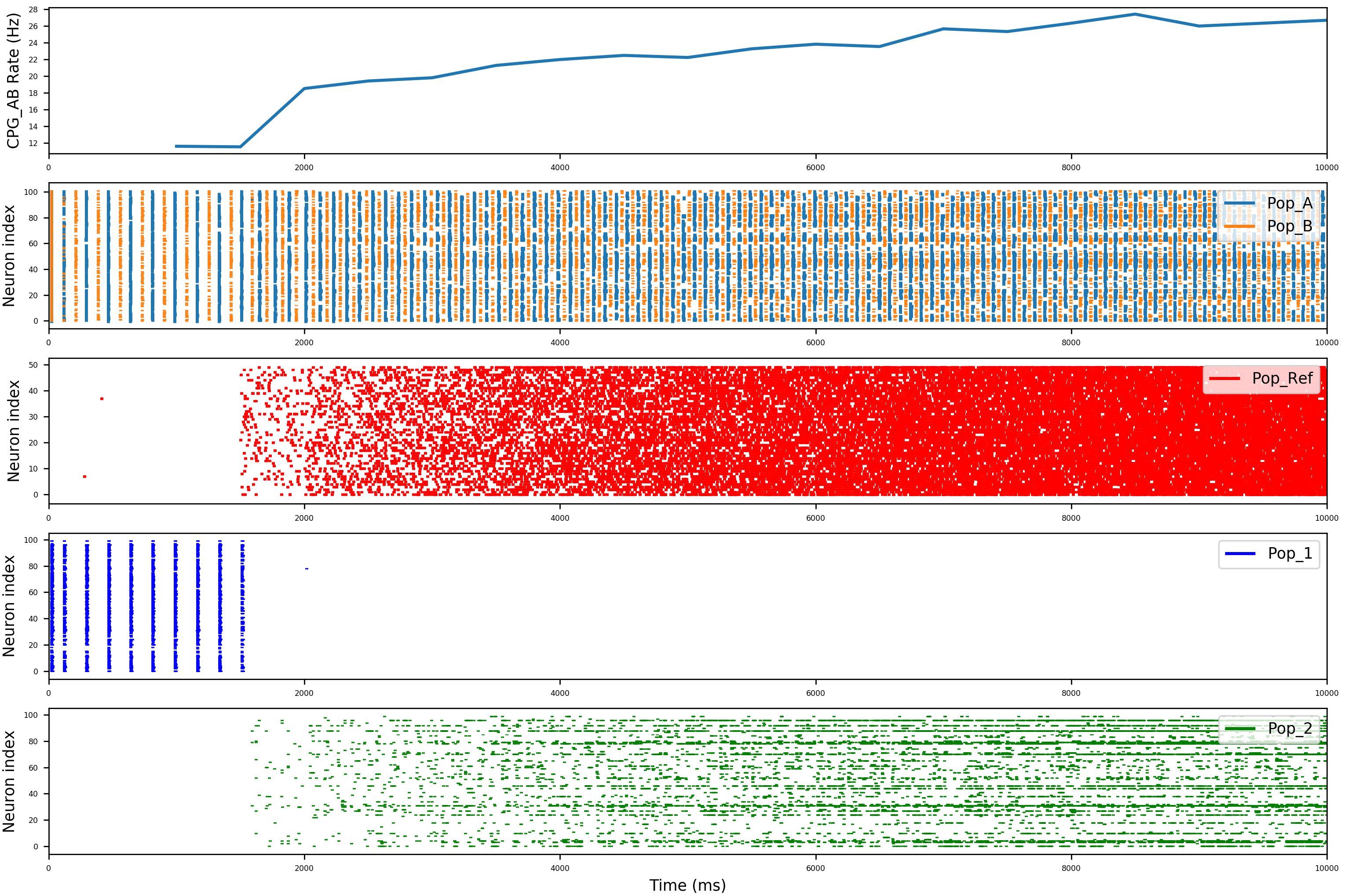}   
    \caption{SpiNNaker simulation of the $CPG$ for ten increasing values of the stimulus rate (from 0~Hz to 50~Hz).}    
   \label{fig:CPGAB_barrido2}
\end{figure}

\subsubsection{$CPG$ under different stimuli}
The SpiNNaker implementation of the full \ac{CPG}, including the feedback network was tested under different conditions of the stimulus. First, we simulated a sudden change of the value of the sensor, represented by the rate of the Poisson generator in the $Ref$ population (see figure \ref{fig:CPGAB_barrido1}). This rate was set to 0.1~Hz for the first 2 seconds of the simulation and oscillated between 171~Hz and 0.1~Hz for the next four seconds, in order to appreciate both regimes of the \ac{CPG}. It can be seen how, with a low $Pref$ frequency, the first feedback population dominates and the rate of the output oscillatory is low, at around 12.5~Hz. With a high $Pref$ frequency, it is the second feedback population which dominates and the measured oscillatory pattern displays a peak frequency of 27.5~Hz. 

\begin{figure}[t]
    \centering
    \includegraphics[width=0.95\textwidth, keepaspectratio]{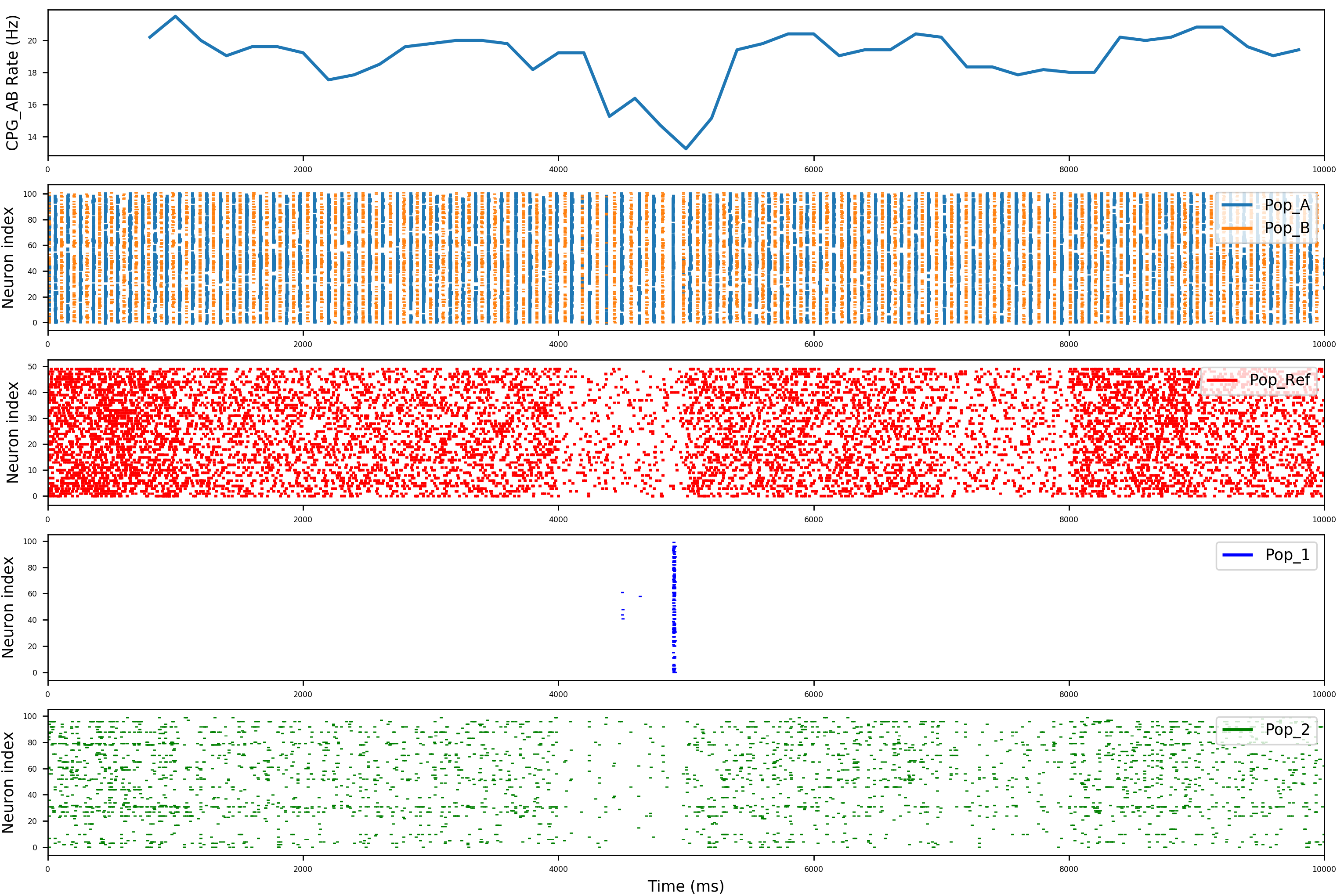}   
    \caption{SpiNNaker simulation of the $CPG$ for ten random values of the stimulus rate (from 0~Hz to 50~Hz).}    
   \label{fig:CPGAB_random}
\end{figure}

Figures \ref{fig:CPGAB_barrido2} and \ref{fig:CPGAB_random} show the spiking response of the SpiNNker \ac{CPG} to increasing rates of $Ref$ and to random values of $Ref$, respectively. The two regimes can be clearly observed in the spiking response of populations~1 and~2 and in the measured oscillatory rates of Populations~A and~B, proving that the feedback mechanism works correctly.

\subsection{Comparison between the results obtained in Brian~2 and SpiNNaker}
\begin{figure}[h!]
    \centering
    \includegraphics[width=0.8\textwidth, keepaspectratio]{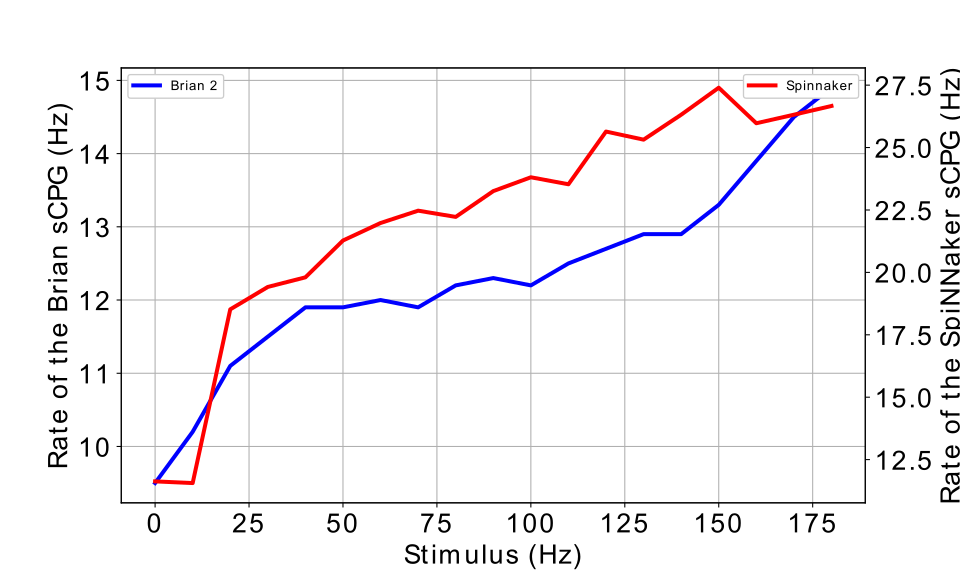}   
    \caption{Comparison of the results obtained in both Brian~2 simulation and SpiNNaker implementation. The plot shows the rate generated by the \ac{sCPG} when the input stimulus (population $Ref$) is changed from 0~Hz to 180~Hz. The blue trace shows Brian~2 results (left y-axis) and the red trace the SpiNNaker results (right y-axis).}    
   \label{fig:comparison}
\end{figure}

Figure~\ref{fig:comparison} shows the comparison made between both approaches: the simulations on Brian~2 and the implementation on the SpiNNaker platform. In this experiment, the same input stimulus from the sensor was used for both approaches, which went from 0~Hz to 180~Hz. The rates generated by both \acp{sCPG} were very similar. In the case of the Brian~2, the operation frequency of the \ac{sCPG} ranged from 9.5~Hz to 14.9~Hz, and from 11.62~Hz to 26.66~Hz in the case of the SpiNNaker platform. The calculated Pearson correlation coefficient between both is 0.905. 

\section{Conclusions}
\label{Sec_Conclusions} 
In this paper, we have presented what, to the best of our knowledge is, the first \Acl{sCPG} that incorporates a feedback in the loop to modify the locomotion frequencies of a legged robot through an adaptive learning mechanism. The feedback was provided by a \ac{FSR} (but any sensor might be used) to simulate the force exerted on each limb of a future robot. With the use of this sensor, we injected different stimuli to the \ac{sCPG}.  Firstly, we performed some tests to find the optimal (minimum) input value, where the feedback effect could be easily observed. Then, some tests were performed to observe the effect in the oscillation frequencies under different stimuli. Finally, we carried out some experiments where the values read from the sensor were alternating between maximum and minimum voltage peaks in order to study the robustness of the \ac{sCPG} against sudden changes in the oscillation frequency. From these experiments, we conclude that our \ac{sCPG} presents a robust behavior in the adaptability of the oscillation frequencies. These frequencies can be used further in the generation of locomotion gaits for legged robots with the advantage that they can be modified depending on the terrain conditions.

The implementation of the \ac{sCPG} was firstly done by using the Brian~2 simulator and then, considering the same parameters, it was migrated to SpiNNaker. The results in both cases are highly the same ones as it is shown in Figure~\ref{fig:comparison}; this fact demonstrates the reproducibility of our architecture on any platform.

Compared to the biological locomotion mechanism, we considered that our approach is highly plausible in two senses, the first one is that the proposed network model is based on spiking neurons, which are considered the neuron models that mimic the best the behavior of biological neurons. The second aspect is that the implementation performed in SpiNNaker allowed us to improve it in terms of the power consumption, and the hardware by itself attempts to be an artificial representation of the brain.

As a future work, we propose to embed the SpiNNaker system into different legged robots (e.g. biped, quadruped and hexapod) to validate our approach on a real robotic platform. Also, it could directly process spatiotemporal patterns with the same \ac{sCPG} by incorporating a neuromorphic sensor.
\section*{Acknowledgements}
This work was partially supported by the Interreg Atlantic Area Programme through the European Regional Development Fund (TIDE - Atlantic network for developing historical maritime tourism, EAPA\_630/2018), by the Spanish grant (with support from the European Regional Development Fund) MIND-ROB (PID2019-105556GB-C33) and by the EU H2020 project CHIST-ERA SMALL (PCI2019-111841-2).






\bibliographystyle{model1-num-names}
\bibliography{references.bib}

\end{document}